\documentclass[sigconf]{acmart}

\usepackage{algorithmic}
\usepackage[linesnumbered,ruled]{algorithm2e}
\usepackage{graphicx}
\usepackage{textcomp}
\usepackage{xcolor}
\usepackage{xspace}
\usepackage{comment}
\usepackage{multirow}
\usepackage{subcaption}
\usepackage{tabularx}

\AtBeginDocument{%
  \providecommand\BibTeX{{%
    \normalfont B\kern-0.5em{\scshape i\kern-0.25em b}\kern-0.8em\TeX}}}

\newcommand\tool{InputReflector\xspace}

\setcopyright{none}
\renewcommand\footnotetextcopyrightpermission[1]{} 
\begin{document}

\title{Generalizing Neural Networks by Reflecting Deviating Data in Production}

%
\author{Yan Xiao}
\affiliation{%
  \institution{School of Computing, \\National University of Singapore}
  \country{Singapore}}
\email{dcsxan@nus.edu.sg}

\author{Yun Lin}
\affiliation{%
	\institution{School of Computing, \\National University of Singapore}
	\country{Singapore}}
\email{dcsliny@nus.edu.sg}

\author{Ivan Beschastnikh}
\affiliation{%
	\institution{Department of Computer Science, \\University of British Columbia}
	\city{Vancouver}
	\state{BC}
	\country{Canada}}
\email{bestchai@cs.ubc.ca}

\author{Changsheng Sun}
\affiliation{%
	\institution{School of Computing, \\National University of Singapore}
	\country{Singapore}}
\email{changsheng\_sun@outlook.com}

\author{David S. Rosenblum}
\affiliation{%
	\institution{Department of Computer Science, \\George Mason University}
	\city{Fairfax}
	\state{VA}
	\country{USA}}
\email{dsr@gmu.edu}

\author{Jin Song Dong}
\affiliation{%
	\institution{School of Computing, \\National University of Singapore}
	\country{Singapore}}
\email{dcsdjs@nus.edu.sg}

%
%
%
%
%
%
%

\begin{abstract}
  Trained with a sufficiently large training and testing dataset, Deep
  Neural Networks (DNNs) are expected to generalize. However, inputs
  may deviate from the training dataset distribution in real deployments. This is a
  fundamental issue with using a finite dataset. Even worse, real
  inputs may change over time from the expected distribution. Taken
  together, these issues may lead deployed DNNs to mis-predict in
  production.

  In this work, we present a runtime approach that mitigates
  DNN mis-predictions caused by the unexpected runtime inputs to
  the DNN. In contrast to previous work that considers the structure
  and parameters of the DNN itself, our approach treats the DNN as a
  blackbox and focuses on the inputs to the DNN. Our approach has two
  steps. First, it recognizes and distinguishes ``unseen''
  semantically-preserving inputs. For this we use a distribution
  analyzer based on the distance metric learned by a Siamese
  network. Second, our approach transforms those unexpected inputs
  into inputs from the training set that are identified as having
  similar semantics. We call this process \emph{input reflection}
  and formulate it as a search problem over the embedding space on the training set.
  This embedding space is learned by a Quadruplet network as an
  auxiliary model for the subject model to improve the generalization.
  

  We implemented a tool called \tool based on the above two-step
  approach and evaluated it with experiments on three DNN models trained
  on CIFAR-10, MNIST, and FMINST image datasets. The results show that \tool
  can effectively distinguish inputs that retain semantics of the
  distribution (e.g., blurred, brightened, contrasted, and zoomed images) and
  out-of-distribution inputs from
  normal inputs.

\end{abstract}




\maketitle

\section{Introduction}
\label{sec:intro}

\begin{figure}[t]
\vspace{4mm}
	\begin{centering}
		\includegraphics[width=0.98\columnwidth]{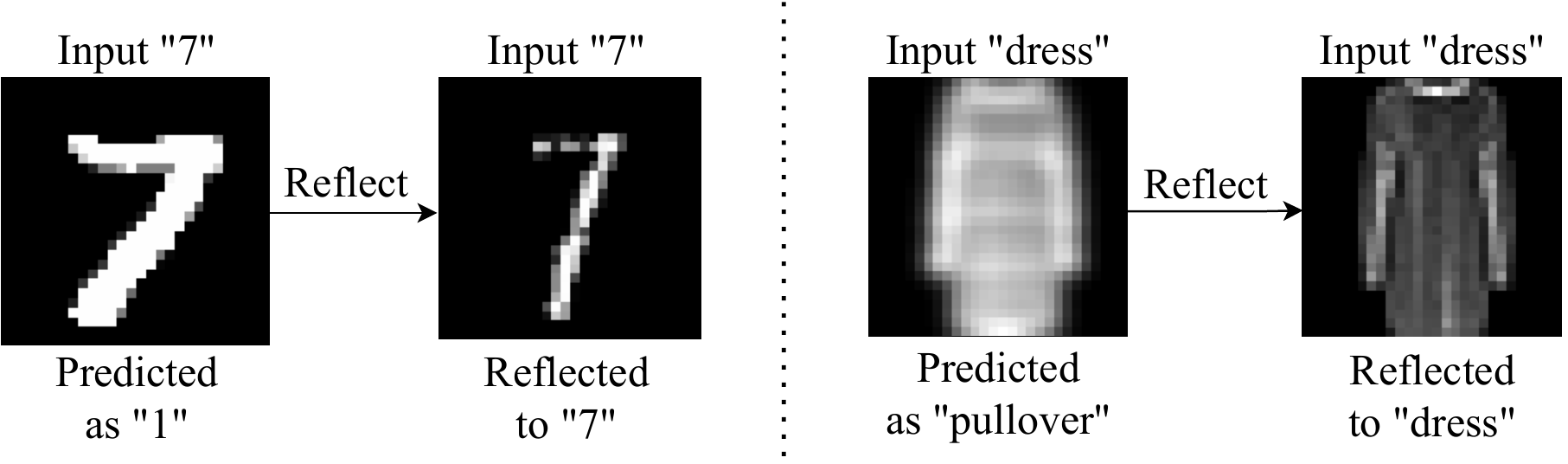}
		\par\end{centering}
	        \caption{Input reflection examples.
                }
	\label{fig:motivate}
	\vspace{-5mm}
\end{figure}

Deep Neural Networks (DNNs) have achieved high performance across many
domains, notably in classification tasks, such as object
detection~\cite{redmon2016you}, image
classification~\cite{he2016deep}, medical
diagnostics~\cite{ciresan2012deep}, and semantic
segmentation~\cite{long2015fully}.  Today, these DNNs are trained on
large amounts of data and then deployed for real-world use.

DNNs require that inputs in
deployment come from the same distribution as the training dataset.
However, real world inputs that are semantically similar to a human observer may
look different from the perspective of the model. For example, the
distribution of data observed by an onboard camera of a self-driving
car may change due to environmental factors; the images may be
brighter or more blurry than those in the training dataset. Even
worse, altogether new and unexpected inputs may be presented to a
deployed model. When presented with such unexpected inputs,
the model may manifest unexpected behaviors. For example, a model
trained on MNIST~\cite{lecun2010mnist} may incorrectly
classify a blurry digit.

%
The academic community and AI industry have proposed a variety of techniques to train an offline robust and trustworthy model, including data augmentation~\cite{gao2020fuzz, samangouei2018defense,zhong2020random,dreossi2018counterexample}, adversarial training~\cite{ganin2016domain,miyato2016adversarial,shafahi2019adversarial,shrivastava2017learning}, and adversarial sample defense techniques~\cite{samangouei2018defense,ilyas2019adversarial,wang2019adversarial,wang2019fakespotter,zhang2020towards}. All of these techniques aim to increase the range of the training dataset so that the trained model can better generalize.
However, this generalization is ultimately constrained by the model architecture and the finite training dataset.
The challenge is that real-world inputs in deployment can have unexpected variations,
and these inputs may be difficult to capture through data augmentation during the training stage.
One reason for this is that data enrichment-based techniques rely on a priori knowledge of the transformations that may appear in deployment.
Furthermore, such solutions may sacrifice accuracy to improve generalization ability.
Recent work considers online retraining/learning approaches~\cite{gao2021wide}, but these have
a high performance cost and may not be appropriate for applications with timing constraints.


For out-of-distribution inputs a variety of techniques exist,
including ODIN~\cite{liang2018enhancing},
Mahalanobis~\cite{lee2018simple}, and Generalized
ODIN~\cite{hsu2020generalized}. These techniques distinguish
out-of-distribution from in-distribution data. But, none of them focuses on identifying deviating data with \textit{similar semantics} 
to the training data. 
That is, they do not
distinguish inputs that are further away from the training dataset but
closer than out-of-distribution data.

We present a runtime input-reflection approach that (1) recognizes
``unseen'' semantic-preserving inputs (e.g., bright or blurry images,
text written by non-native speakers with grammatical deviations,
dialect speech inputs)
for a DNN model, and (2) reflects them to samples in the training
dataset that conform to a similar distribution. This second step
improves model generalization. The key insight of our approach is that
an ``unseen'' input can be potentially mapped into a semantically
similar ``seen'' input on which the model can make a more trustworthy
decision.

Figure~\ref{fig:motivate} presents two examples where the input
reflection process corrects the classification of a brightened digit
``7'' and a blurry ``dress''. The digit ``7'' presents a different
handwriting style (e.g., with thicker strokes than those in the
training data), causing the model to miss-predict it as a ``1''.  Our
reflection approach , without learning any handwriting style, can
reflect this input into a digit ``7'' from the training dataset to
mitigate the miss-prediction. Similarly, the right side of
Figure~\ref{fig:motivate} shows a blurry ``dress'' input.  The DNN
model incorrectly classifies this input as a ``pullover''. By
contrast, the reflected version is correctly classified as a
``dress''.  Both examples illustrate how input reflection can help
DNNs deal with unexpected inputs in production.

Realizing input reflection requires addressing two technical
challenges.  First, we need a technique to distinguish
normal inputs (e.g., a digit from the MNIST dataset), semantically
similar inputs unseen during training (e.g., a blurry or bright
digit), and out-of-distribution inputs (e.g., a non-digit image).  The
second challenge is to design a reflection process to map
inputs unseen during training
into semantically similar inputs from the training dataset.

There are a variety of DNN models. In this work we make the first
attempt to reflect inputs by overcoming the above challenges in the
context of computer vision models. We leave other domains for future
work.  For challenge (1), we invent a measurement based on a
Siamese network, which can discriminate
the normal input, unseen deviated input, and out-of-distribution
input.  For challenge (2), we convert the reflection
problem into a problem of finding similar inputs in the training
dataset based on the learned embeddings by a Quadruplet network, which
can capture the similarity of learned embedding from two images.

In summary, we make the following contributions:
\begin{itemize}

\item We propose a novel technique to improve DNN generalization by
  focusing on DNN inputs and treating the DNN as a black-box in production.
  %
  %
  Our technique contributes (1) a general
  discriminative measurement to distinguish normal, deviated, and
  out-of-distribution inputs; and (2) a search method to reflect an
  unexpected input into one that has similar semantics and is present
  in the training set.

\item We implemented a tool called \tool based on the above approach,
  and evaluate it on three DNN models (ConvNet, VGG-16, and ResNet-20)
  trained on the CIFAR-10, MNIST, and FMNIST datasets. Our experimental
  results show that our tool \tool can recognize 77.19\% of the
  deviated inputs and fix 77.50\% of them. 
  Our implementation is open-source\footnote{https://github.com/yanxiao6/InputReflector-release}.



\end{itemize}

\section{Background}
\label{sec:background}


In this section we review Siamese
networks and triplet loss, which are key to our solution.

%


\subsection{Siamese Network}

Bromley et al.\cite{bromley1993signature} introduced the idea of a
Siamese network to quantify the similarity between two images. Their
Siamese network quantifies the similarity between handwriting
signatures. Typically, a Siamese network transforms the inputs
$\textbf x$ into a feature space $\textbf z = f(\textbf x)$, where
similar inputs have a shorter distance and dissimilar inputs have
larger distance.  Given a pair of inputs $\textbf x_1$ and $\textbf
x_2$, we usually use Cosine similarity or Euclidean distance to
compare $f(\textbf x_1)$ and $f(\textbf x_2)$. Siamese
networks~\cite{chicco2021siamese} have been applied to a wide-range of
applications, like object
tracking~\cite{zhang2019deeper,li2019siamrpn++}, face
recognition~\cite{taigman2014deepface}, and image
recognition~\cite{koch2015siamese}. In this work, we use Siamese
networks to evaluate the semantic similarity between an input in
deployment and samples in the training dataset.



\subsection{Triplet Loss}

\begin{figure}[t]
	\begin{centering}
		\includegraphics[width=0.9\columnwidth]{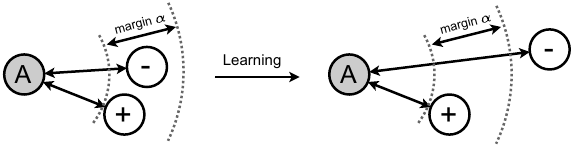}
		\par\end{centering}
	\caption{An example of using triplet loss.}
	\label{fig:triplet}
\end{figure}

Triplet loss was first proposed in FaceNet~\cite{schroff2015facenet}.
Given a training dataset where samples are labeled with classes, the
triplet loss is designed to project the samples into a feature space
where samples under the same class have shorter distance than those
under different classes. \autoref{fig:triplet} shows an example of
triplet loss.  Given a sample as \textit{Anchor} of class $c_1$,
taking a positive sample (annotated as ``+'') under $c_1$ and a
negative sample (annotated as ``-'') under $c_2$, triplet loss is
designed to pull the positive sample closer while pushing the negative
sample further from the anchor. Technically, the definition of
triplet loss is as follows:
%
%
%
%
\begin{equation}\label{eq:triplet}
	\mathcal{L}_\text{triplet} = max(Dis(Anchor, Pos) - Dis(Anchor, Neg) + \alpha, 0)
\end{equation}

In the above, $\alpha$ is the margin, introduced to keep a
distance gap between positive and negative samples.  The loss can be
optimized if the Siamese network can push $Dis(Anchor, Pos)$ to 0 and
$Dis(Anchor, Neg)$ to be larger than $Dis(Anchor, Pos) + \alpha$.

Next, we describe our approach, which uses Siamese
networks and triplet loss to realize input reflection.



\begin{figure}[t]
	\begin{centering}
		\includegraphics[width=0.95\linewidth]{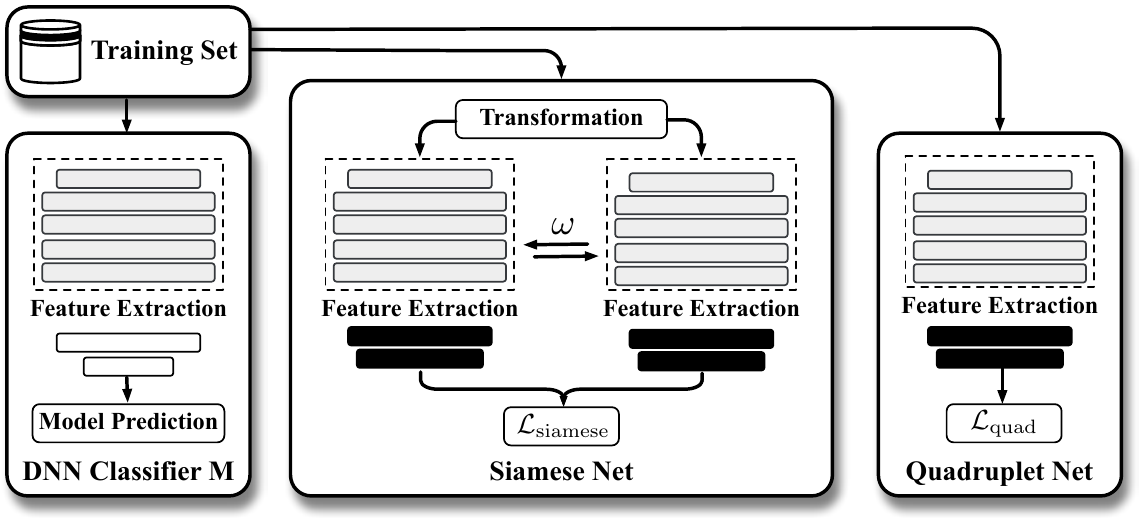}
		\par\end{centering}
	\caption{The design of \tool in training.}
	\label{fig:framework_train}
\end{figure}

\begin{figure}[t]
	\begin{centering}
		\includegraphics[width=0.98\linewidth]{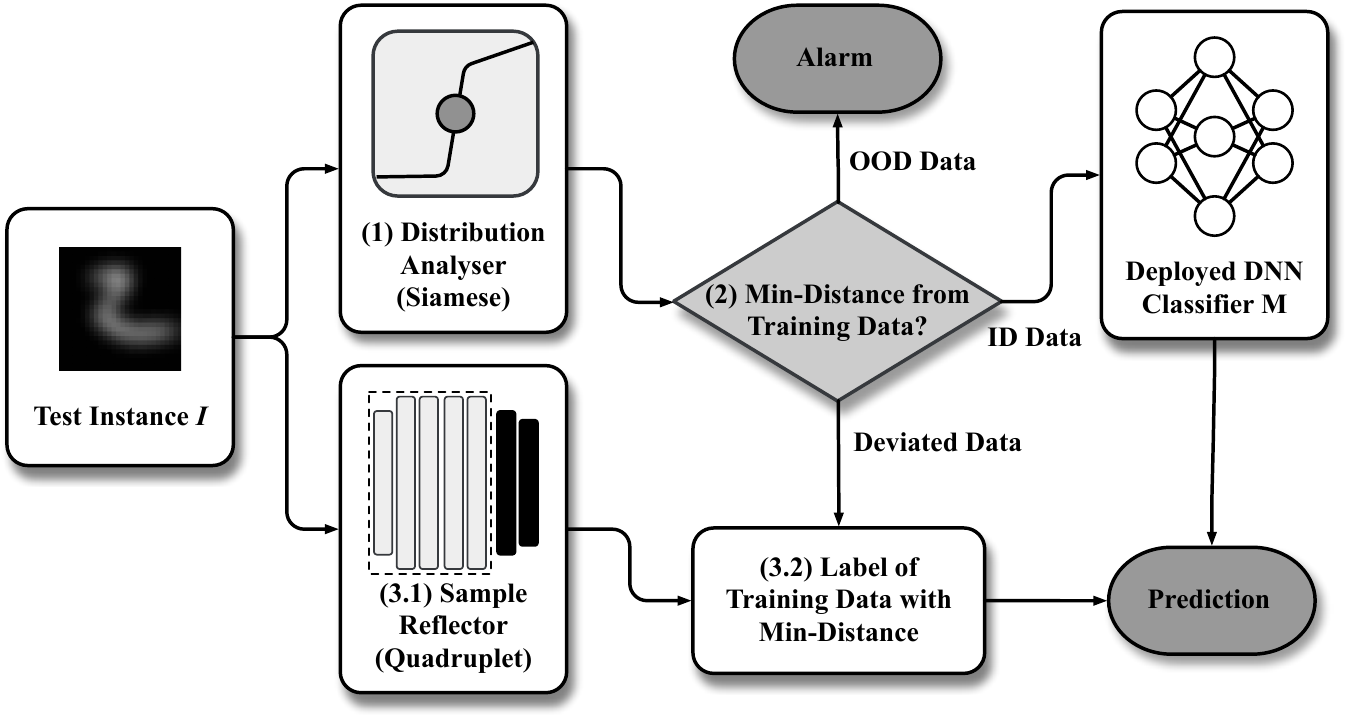}
		\par\end{centering}
	\caption{The design of \tool in deployment.}
	\label{fig:framework_deploy}
\end{figure}

\section{Design of InputReflector}\label{sec:design}


\autoref{fig:framework_train} and \autoref{fig:framework_deploy} reviews the training-time and runtime design
of \tool. We will refer to these figures in this section.

\tool uses information during model training to inform its runtime
strategy. We review how \tool acquires the required information during
training later in this section. At \emph{runtime}, given an input $I$ and a
deployed model $\mathit{M}$, \tool follows a three-step approach:

\begin{enumerate}

  \item Determine if $\mathit{M}$ can correctly predict $I$ by
    checking how well $I$ conforms to the distribution of training
    dataset.
  
  \item If $\mathit{M}$ cannot correctly predict $I$, then determine
    if $I$ shares similar semantics to the training samples.
  
  \item If $I$ is semantically similar to the training samples, then
    determine which training sample is best used instead of $I$ for
    the prediction.

\end{enumerate}

The key to the above steps is correct characterization of the
semantics of an input. For this, we rely on sample distribution and
design a \textbf{Distribution Analyzer} (Section~\ref{sec:analyzer}).
This piece of our approach is inspired by prior
research~\cite{qian2019softtriple,kim2020proxy} that indicates that the
distribution of embedding vectors can be used to group semantically
similar data (e.g., images of the same class). \tool uses
the embedding distribution to check the semantic similarity
between the test instance and the training data.
For inputs that are classified by the Distribution Analyzer as
deviating samples, we design the \textbf{Sample Reflector}
(Section~\ref{sec:selector}) to search for a close training sample to
replace the runtime test instance. 

We designed the Distribution Analyzer and Sample Reflector as auxiliary
DNNs, for their inherent expressiveness. The Distribution Analyzer,
implemented as a Siamese network, captures the general landscape of
in-distribution, deviated, and out-of-distribution samples. The Sample
Reflector, implemented as a Quadruplet network, calculates the detailed
distance measurement between the samples.

Next, we detail the Distribution Analyzer and Sample Reflector.

\subsection{Distribution Analyzer design}\label{sec:analyzer}

A challenge in building an effective Distribution Analyzer is
\emph{designing a smooth measure for semantically same, similar, and
  dissimilar samples.}  Specifically, we need to design the auxiliary model so
that samples can be projected into a space where there is such
smoothness between the three categories.

To address this challenge, we require the auxiliary model to learn
the smooth measurement from in-distribution, deviated, and
out-of-distribution samples.  Specifically, let an in-distribution
sample be $x$, $K_{in}$ be the distance between $x$ and its most
different semantic-preserving sample $x_s$, $K_{out}$ be the distance
between $x$ and its most similar semantic-different sample $x_o$, any
of semantic-preserving and deviating sample $x'$ should conform to:
\begin{equation}\label{eq:dist}
	K_{in} < min_{x_{t} \in \mathcal{X}^{t}}(f(x', x^{t})) < K_{out}
\end{equation}

To train a quality auxiliary model we prepare \textit{representative}
transformed samples that preserve the semantics of the training data.
Specifically, our auxiliary models are trained on three types of data:
(1) in-distribution samples (i.e., original training data),
(2) deviating samples (i.e., representative transformed data), and
(3) one kind of out-of-distribution samples (i.e., extremely transformed data, discussed in Section~\ref{sec:setup}).

Next we discuss how we design the similarity function $f$ and how we generate $K_{in}$ and $K_{out}$.

\subsubsection{Siamese Network Training}
\begin{figure}[t]
	\begin{centering}
		\includegraphics[width=0.85\linewidth]{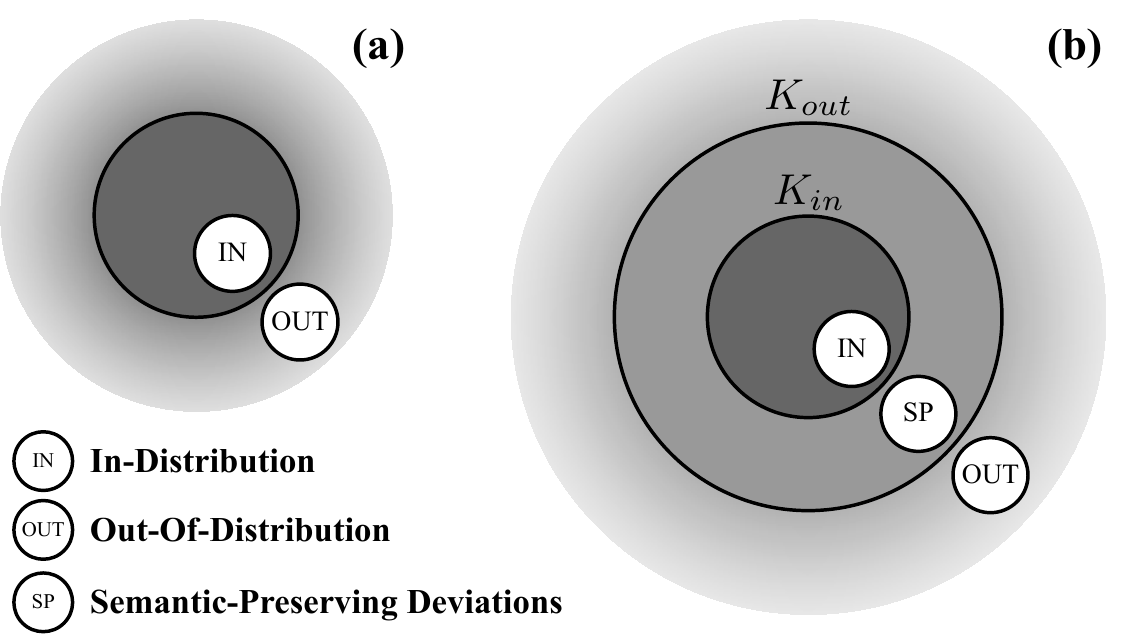}
		\par\end{centering}
	\caption{Comparison of problem formulation and thresholds used
          by (a) previous work, and (b) this paper.}
	\label{fig:threshold}
\end{figure}

Existing work on detecting out-of-distribution data
uses one threshold Figure~\ref{fig:threshold}(a). By contrast,
the challenge in our work is to find two thresholds,
$K_{in}$ and $K_{out}$ in Figure~\ref{fig:threshold}(b), so that
we can discriminate three types of data. Inspired by the Siamese
network with triplet loss, which are used to identify people across
cameras~\cite{schroff2015facenet,chen2017beyond}, we build a Siamese
network to learn two thresholds ($K_{in}$ and $K_{out}$) so that we
can discriminate the three kinds of data.

To split the three kinds of data, both training data and their
transformed versions are needed. For example, if we consider the blur
transformation as an example, then the three kinds of data would be
the normal training data $x$, blurry training data $x'$, and very
blurry training data $x''$ (Figure \ref{fig:framework_train}).

Our goal is to learn feature embeddings from the training dataset that
push the semantic-preserving deviating input away from both the
in-distribution data and out-of-distribution data. Specifically, the
Siamese network in the distribution analyzer is designed to make the
distance between the very blurry training data and the normal data
larger than the distance between the blurry training data and the
normal data. Formally, the aim of the Siamese network is $f(x_{c_{i}},
x_{c_{j}}) + K_{in} < f(x_{c_{i}}, x^{'}_{c}) < f(x_{c_{i}},
x_{c_{j}}) + K_{out} < f(x_{c_{i}}, x^{''}_{c})$. The loss function of this network
is designed to minimize the objective:

\begin{equation} \label{eq:loss}
	\begin{aligned}
	L=&max(f(x_{c_{i}}, x_{c_{j}}) - f(x_{c_{i}}, x^{'}_{c}) + m_{1}, 0) + \\
	&max(f(x_{c_{i}}, x^{'}_{c}) - f(x_{c_{i}}, x_{c_{j}}) - m_{2}, 0) + \\
	&max(f(x_{c_{i}}, x_{c_{j}}) - f(x_{c_{i}}, x^{''}_{c}) + m_{2}, 0)
	\end{aligned}
\end{equation}
where $m_{1}$ and $m_{2}$ are the values of margins and $m_{1} <
m_{2}$, $x_{c_{i}}$, $x_{c_{j}}$, and $x_{c}$ denote instances with
the same label $c$ but $i \neq j$ which means that they are different
instances.

This loss function consists of three components. The first and the
last max components push both transformed versions of the training
data away from the normal data but keep the extremely transformed
version further away using a larger margin $m_{2}$. The middle max
component is designed to distinguish transformed ($x'$) and extremely
transformed ones ($x''$).

\begin{figure}[t]
	\begin{centering}
		\includegraphics[width=0.85\linewidth]{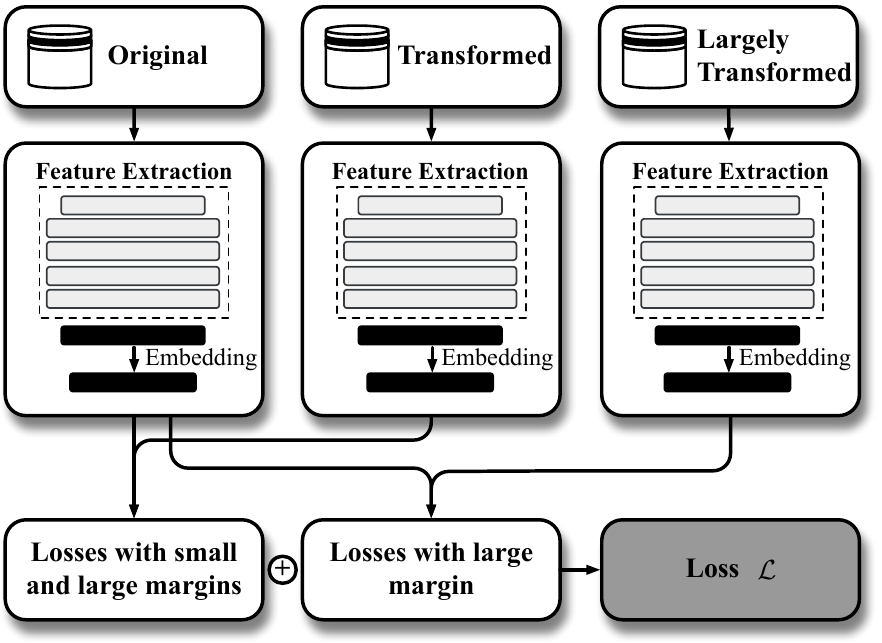}
		\par\end{centering}
	\caption{The architecture of the Siamese network used during the training phase.}
	\label{fig:siamese}
\end{figure}

Figure~\ref{fig:siamese} shows the architecture of our Siamese
network. Three kinds of data (original training data, transformed
training data, and largely transformed training data) are given to the
three models that share weights with each other. During the original
classification training, the DNN classifier $M$ learns to extract
meaningful and complex features in feature extraction layers before
these are fed to the classification layers. To benefit from these rich
representations of feature extraction layers, the Siamese network
builds on $M$ to pass these features through a succession of dense
layers. The outputs of the final dense layer are embeddings that will
be used to minimize the loss function in equation~\ref{eq:loss}. The
function of the similarity measurement $f$ is the distance between the
embeddings of two instances, formally, $f(a, b)=\big\lVert ebd(a) -
ebd(b)\big\rVert_2^2$.

\subsubsection{Distribution Discrimination}

%
We designed the distribution analyzer to distinguish between the
three kinds of data. 
%
In the training module, the embeddings of the original training data
and validation data are generated from the trained Siamese
network. For each validation data, its embedding is used to search for the
closest one from the training data by the distance measure
$f$. $\mathit{K}_{in}$ and $\mathit{K}_{out}$ are selected from
the distances of embeddings between validation and training data.
In deployment, given a test instance $x$, the minimum distance between its
embedding and those of the training data will be a discriminator to
determine its data category:
\begin{equation}
	\left\{\begin{array}{ll}
		\text{In-distribution data}            &  {min_{x_{t} \in \mathcal{X}^{t}}(f(x, x^{t})) < \mathit{K}_{in}}\\
		\text{Out-of-distribution data}    &  {min_{x_{t} \in \mathcal{X}^{t}}(f(x, x^{t})) > \mathit{K}_{out}}\\
		\text{Deviated data}                    &  {others}
	\end{array}\right.
\end{equation}

%
Note that unlike other enhanced classifier approaches~\cite{wang2020dissector,xiao2021self},
\tool understands potential model prediction risks and can be made to
discard runtime test instances that are unsuitable for model prediction
(the \textbf{Out-of-Distribution Alarm} in Figure \autoref{fig:framework_deploy}).

\subsection{Sample Reflector design}\label{sec:selector}

When the distribution analyzer recognizes an input instance as a
deviation from the in-distribution data, the \emph{sample reflector} will map
this input to the closest sample in the training data. To build the
sample reflector we adopt the idea of separation of
concern~\cite{pham2008prediction}. That is, the subject model just
needs to fit a limited number of samples in the training dataset,
while the auxiliary model tackles the similarity measurement between
training samples and their deviations. In other words, we let the
subject model focus on fitting, and we task the auxiliary models with
\textit{generalizability learning}.

Next, we discuss the design of the sample reflector.


\subsubsection{Quadruplet Network Training}

Inspired by the work of Chen et al.~\cite{chen2017beyond} who built a
deep quadruplet network for person re-iden\-tifi\-cation, we construct a
quadruplet network using the subject DNN classifier $M$ (similar to
our Siamese network) to pull instances with the same label closer and
push away instances with different labels. As illustrated in
Figure~\ref{fig:framework_train}, the goal of the Quadruplet network is to
learn the feature embeddings so that instances of intra-class will be
clustered together but those of inter-class will be further away. As a
result, when an unexpected instance is presented, its embedding can be
used to search for the closest samples from the training data, and
these can be used as a proxy for identifying the input class.

\begin{algorithm}[t]
	
	\caption{Quadruplet Network Training and Inference}
	
	\label{alg:quad}
	\KwIn{Target DNN classifier: $M$\;
		Input instances in $\mathit{D_{train}}$: $\mathcal{X}^{t}$, true labels in $\mathit{D_{train}}$: $\mathcal{Y}^{t}$\;
		Deviated test instance: $\mathcal{X}^{d}$}
	\KwOut{Trained Quadruplet network\;
		Alternative predictions of the transformed test instance}
	\# Training\\
	$base\_model = M.get\_layer(``cnn").output$\;
	Add several dense layers after $base\_model$ to be embeddings $embed$\;
	$model = Model(base\_model.input, embed)$\;
	$model.compile(optimizer, Quadruplet\_loss)$\;
	$model.fit(\mathcal{X}^{t}, \mathcal{Y}^{t})$\;
	
	\# Inference\\
	$embed^{t}=model.predict(\mathcal{X}^{t})$\;
	\For{$\mathbf{x}$ in $\mathcal{X}^{d}$}
	{
		$embed(x)=model.predict(x)$\;
		$idx^{T} = argmin(dist(embed(x), embed^{t}))$\;
		$predction_{new} = \mathcal{Y}^{T}[idx^{T} ]$\;
	}
\end{algorithm}

Algorithm~\ref{alg:quad} lists the details of our Quadruplet network
construction and how we use it in deployment. The Quadruplet network
consist of the feature extraction layers of the target DNN classifier
$M$ (line 2) and a succession of dense layers as shown in lines 2-4,
which are trained by the quadruplet loss that is presented in
Algorithm~\ref{alg:loss}.  This is an improvement over the triplet
loss~\cite{schroff2015facenet} with an online component for mining the
triplet online~\cite{hermans2017defense}.

During the construction of the Quadruplet network, the challenge is
how to sample the quadruplet from the training
data. Algorithm~\ref{alg:loss} discusses how to mine quadruplet
samples during training. The sample mining process of the Siamese
network in Section~\ref{sec:analyzer} also uses this technique.

\begin{algorithm}[t]
	
	\caption{Sampling in Quadruplet\_loss}
	
	\label{alg:loss}
	\KwIn{True labels in $\mathit{D_{train}}$: $\mathcal{Y}^{t}$\;
		Margins: $m_{1}$ and $m_{2}$\;
		Embeddings: $embed$}
	\KwOut{loss}
		$pairwise\_dist = dist(embed)$\;
		obtain $mask$ where $mask[a, p]$ is True if $a$ and $p$ are distinct and have same label\;
		$dist_{ap} = mask * pairwise\_dist$\;
		$dist_{hard_{p}} = max(dist_{ap}, axis=1)$\;
		obtain $mask$ where $mask[a, n]$ is True if $a$ and $n$ have distinct labels\;
		$max\_dist_{an} = max(pairwise\_dist, axis=1)$\;
		$dist_{an} = pairwise\_dist + max\_dist_{an} * (1.0 - mask)$\;
		$dist_{hard_{n}} = min(dist_{an} , axis=1)$\;
		$loss_{an} = mean(max(dist_{hard_{p}} - dist_{hard_{n}}  + m_{1}, 0.0))$\;
		\vspace{1em}
		obtain $mask$ where $mask[a, p, n]$ is True if the triplet $(a, p, n)$ is valid\;
		$max\_dist_{an} = max(pairwise\_dist)$\;
		$dist_{an} = pairwise\_dist + max\_dist_{an} * (1.0 - mask)$\;
		$dist_{hard_{nn}} = min(min(dist_{an} , axis=2), axis=1)$\;
		$loss_{nn} = mean(max(dist_{hard_{p}} - dist_{hard_{nn}}  + m_{2}, 0.0))$\;
		$loss = loss_{an} + loss_{nn}$\;
	
\end{algorithm}

The loss of the Quadruplet network consists of two parts (Line 15 in
Algorithm~\ref{alg:quad}). The first part, $loss_{an}$, is the
traditional triplet loss that is the main constraint. The second part,
$loss_{nn}$, is auxiliary to the first loss and conforms to the
structure of traditional triplet loss but has different triplets. We use two
different margins ($m_{1} > m_{2}$) to balance the two
constraints. We now discuss how to mine triplets for each loss.

First, a 2D matrix of distances between all the embeddings is
calculated and stored in $pairwise\_dist$ (line 1). Given an anchor,
we define the \emph{hardest positive example} as having the same label
as the anchor and whose distance from the anchor is the largest
($dist_{hard_{p}}$) among all the positive examples (lines
2-4). Similarly, the \emph{hardest negative example} has a different
label than the anchor and has the smallest distance from the anchor
($dist_{hard_{n}}$) among all the negative examples (lines 5-8).
These \emph{hardest positive example} and \emph{hardest negative
  example} along with the anchor are formed as a triplet to minimizing
$loss_{an}$ in line 9. After convergence, the maximum intra-class
distance is required to be smaller than the minimum inter-class
distance with respect to the same anchor. 
%

To push away negative pairs from positive pairs, we introduce one more loss, $loss_{nn}$.
Its aim is to make the maximum intra-class distance
smaller than the minimum inter-class distance regardless of whether
pairs contain the same anchor.
%
%
This loss constrains the distance between positive pairs (i.e.,
samples with the same label) to be less than any other negative pairs
(i.e., samples with different labels that are also different from the
label of the corresponding positive samples). With the help of this
constraint, the maximum intra-class distance must be less
than the minimum inter-class distance regardless of whether pairs
contain the same anchor. To mine such triplets, the valid triplet first
needs to be filtered out on line 10 where $i, j, k$ are distinct and
$(labels[i] \neq labels[j]) \ \& \ (labels[i] \neq labels[k]) \ \& \ (labels[j] \neq labels[k])$.
Then, the hardest negative pairs are sampled whose distance is the
minimum among all negative pairs in each batch during training (line
11-13).  Finally $loss_{an}$ is minimized to further enlarge the
inter-class variations in Line 14.

As discussed in~\cite{chen2017beyond}, $(loss_{an} + loss_{nn})$
leads to a larger inter-class variation and a smaller intra-class variation
as compared to the triplet loss. And this loss combination has a better generalization
ability as will show in the evaluation (Section~\ref{sec:eval}). The benefit
of this arrangement is that the Quadruplet network can be used to help
generalize the subject classifier $M$ to the unexpected deviations from
in-distribution data.

\subsubsection{Input Reflector for unexpected instances}

Algorithm~\ref{alg:quad} lists the procedure for input
reflection. Given an unexpected instance $x$ that belongs to the
variated in-distribution data in deployment, the algorithm first
obtains its feature embedding learned by the Quadruplet network (line
10). Next, the distance between $embed(x)$ and those embeddings of the
training data generated in the training module (line 8) will be
calculated so that the minimum distance can be found.

Finally, the unexpected test instance $x$ is then reflected to the
specific training instance with the minimum distance whose label
becomes the alternative prediction for $x$ (lines 11-12).

Now that we have detailed \tool's design, we evaluate its performance
on different datasets and DL models.

\section{Evaluation}
\label{sec:eval}

We now present experimental evidence for the effectiveness of \tool. Our evaluation goal is to answer the following research questions:


\noindent \textbf{RQ1. Distribution Analyzer:} \emph{How effective is the distribution analyzer in distinguishing three types of data?}
To evaluate the distribution analyzer, we compare its area under the
receiver operating characteristic curve (AUROC) with related
techniques, namely, Generalized ODIN (G-ODIN)
\cite{hsu2020generalized} and SelfChecker
\cite{xiao2021self}. Specifically, we evaluate the performance of
the distribution analyzer on detecting out-of-distribution and deviated
data.  Since the authors of G-ODIN did not release the code nor
specific values for the hyperparameters, we implemented
G-ODIN ourselves and used grid search to find the best
combination of hyperparameters on the validation dataset, from which we
obtained close results on the dataset in the G-ODIN
paper. As there are two parts to SelfChecker, alarm and
advice, to answer this RQ, we compare the distribution analyzer
with SelfChecker's alarm accuracy.


\vspace{1em}
\noindent \textbf{RQ2. Sample Reflector Accuracy:} \emph{What is the accuracy of the reflection process on unseen deviated data?}
In cases where \tool recognizes the test instance as a deviation from
in-distribution data, the reflection process will be used to 
provide an alternative prediction. To answer this question, we compare
the accuracy of sample reflector (Section \ref{sec:selector}) in \tool
against the accuracy of the subject model $M$ and advice accuracy of
SelfChecker on the unseen deviated input.


\vspace{1em}
\noindent \textbf{RQ3. \tool Performance:} \emph{What is the performance of \tool on the in-distribution and deviated input in deployment?}
Since the distribution analyzer is designed to learn $K_{in}$ and
$K_{out}$ from transformed versions of training data (blur in this
work), we compare the accuracy of \tool against the accuracy of
original subject model ($M$), subject model with data augmentation
($M+Aug$), and SelfChecker on both pure in-distribution inputs and the
deviating inputs.

\vspace{1em}
\noindent \textbf{RQ4. Overhead:} \emph{What is the time/budget overhead of \tool in both training and deployment?}
To evaluate the efficiency of \tool, we calculated its overall time
overhead with the ones of $M+Aug$ and SelfChecker. In the training
phase, $M+Aug$ needs to conduct data generation and extra
training. But, we regard the generation part as having zero cost and
only consider the extra training time. Augmentation also does not have
extra time cost in deployment. For SelfChecker and \tool, we measure their
time overheads in training and deployment.

\subsection{Experimental Setup}\label{sec:setup}

We evaluate the performance of \tool on three popular image datasets
(MNIST~\cite{lecun2010mnist}, FMNIST~\cite{xiao2017fashion}, and
CIFAR-10~\cite{krizhevsky2009learning}) using three DL models
(ConvNet~\cite{kim2019guiding}, VGG-16~\cite{simonyan2014very}, and
ResNet-20~\cite{he2016deep}).  We conducted all experiments on a Linux
server with Intel i9-10900X (10-core) CPU @ 3.70GHz, one RTX 2070
SUPER GPU, and 64GB RAM, running Ubuntu 18.04.


Our evaluation focuses on computer vision models. We hope to consider
other types of models in future work.
To evaluate \tool, we prepare three kinds of image datasets as follows:

\vspace{.2em}
\noindent \textbf{In-distribution datasets.}
In-distribution data conform to the
distribution of the training data. As in prior studies
\cite{hsu2020generalized, liang2018enhancing, lee2018simple}, we
regard the testing data of each dataset as the in-distribution data.

MNIST is a dataset for handwritten digit image recognition, containing
60,000 training images and 10,000 test images, with a total number of
70,000 images in 10 classes (the digits 0 through 9).  Each MNIST
image is a single-channel of size 28 * 28 * 1.  Similarly,
Fashion-MNIST is a dataset consisting of a training set of 60,000
images and a test set of 10,000 images. Each image is 28 * 28
grayscale images associated with a label from 10 classes.  CIFAR-10 is
a collection of images for general-purpose image classification,
including 50,000 training data and 10,000 test data with 10 different
classes (airplanes, cars, birds, cats, etc.). Each CIFAR-10 image is
three-channel of size 32 * 32 * 3. The classification task of CIFAR-10
is generally harder than that of MNIST due to the size and complexity
of the images in CIFAR-10.

\vspace{.2em}
\noindent \textbf{Deviations from in-distribution datasets.}
We selected four kinds of transformations to construct the transformed
data of in-distribution testing datasets: blur, bright, contrast, and
zoom. These operations transform an image $x$ as follows
\cite{gao2020fuzz}:

\begin{itemize}

\item $zoom(x,d)$: zoom in $x$ with a zoom degree $d$ in range [1,5).
  
\item $blur(x,d)$: blur $x$ using Gaussian kernel with a degree $d$
  in range [0,5).
  
\item $bright(x,d)$: uniformly add a value for each pixel of $x$
  with a degree $d$ in range [0, 255) and then clip $x$ within [0,
      255].
  
  \item $contrast(x,d)$: scale the RGB value of each pixel of $x$ by a
    degree $d$ in range (0, 1] and then clip $x$ within [0, 255].
  
\end{itemize}

We search for crash transformation degrees for each training and
testing image on which the original classifier begins to
mis-predict. The instances with such degrees then serve as the deviations
of in-distribution data.

\vspace{.2em}
\noindent \textbf{Out-of-distribution datasets.}
There are two kind of out-of-distri\-bution data: the first one is the
same as the existing work --- another dataset which is completely
different from the training data. The second one is the extremely
transformed training and testing data using extreme
degrees of the four transformations.

Specifically, we use FMNIST, MNIST, SVHN~\cite{netzer2011reading} as the first totally
different dataset for MNIST, FMNIST, CIFAR-10 respectively. The
extremely transformed data cannot be recognized by both DL models and
humans. We present the performance of the distribution analyzer on
both dataset: another dataset and extremely transformed testing data.

\vspace{.2em}
\noindent \textbf{DL models.}
We choose three DL models as our subject models: ConvNet, VGG-16, and
ResNet-20. These are commonly-used models whose sizes range from
small to large, with the number of layers ranging from 9 to 20.

\vspace{.2em}
\noindent \textbf{Configurations.}
For fair comparison, the subject model $M$ does not use data
augmentation. As discussed in Section \ref{sec:analyzer} and
\ref{sec:selector}, both the Siamese and Quadruplet networks are built
on the subject model $M$, along with a succession of dense
layers. We set the number of dense layers to be the same as the number of
the classification layer in $M$. We use the Euclidean distance 
as the distance metric $f$ in (\ref{eq:dist}).

\vspace{.2em}
\noindent \textbf{Evaluation metrics.}
We adopt two measures to evaluate the distribution analyzer. AUROC
plots the true positive rate (TPR) against the false positive rate
(FPR) by varying a threshold, which can be regarded as an averaged
score that can be interpreted as the model's ability to discriminate
between positive and negative inputs.  TNR@TPR95 is the true negative
rate at 95\% true positive rate, which simulates an application
requirement that the recall should be 95\%. Having a high TNR under a
high TPR is much more challenging than having a high AUROC score.
We use classical model accuracy on testing data to evaluate the
performance of both sample selector and \tool.

\subsection{Results and Analyses}\label{sec:results}

We now present results that answer our four research questions.

\vspace*{0.2em}\noindent\textbf{RQ1. Distribution Analyzer.}\vspace*{0.2em}
Table \ref{tab:sia_out} presents the AUROC and TNR@TPR95 of the
distribution analyzer (Siamese) and Generalized ODIN (G-ODIN) to
detect out-of-distribution data from in-distribution and deviated
data. We exclude SelfChecker here because of its assumption on
handling input that share similar semantics with the training
dataset. As mentioned in Section \ref{sec:setup}, the
out-of-distribution data here consists of totally differently data
(same to the existing work) and extremely transformed data (blur,
bright, contrast, and zoom), that's why the results in Table
\ref{tab:sia_out} are different from previous work on detecting
out-of-distribution from in-distribution data. Intuitively,
distinguishing between the three data types is more difficult.

Based on Table \ref{tab:sia_out}, Siamese outperforms G-ODIN in most
of the cases except for FMNIST on VGG-16. And we can see that both
techniques have similar AUROC on FMNIST. The main reason is that
Siamese cannot distinguish well between the deviated data and
out-of-distribution data (MNIST here). When FMNIST data is contrasted,
e.g., trouser, it will be similar to digit "1". For other settings,
Siamese can not only detect totally different data as G-ODIN but also
detect more extremely transformed data on which G-ODIN has bad
performance. In average, Siamese achieves 80.52\% AUROC against 71.02\% of G-ODIN.

\begin{table*}[h]
	\centering
	\caption{Performance of two methods on detecting out-of-distribution data.}
	\label{tab:sia_out}
    \begin{tabular}{lllllll}
    \toprule[0.25ex]
    \multicolumn{1}{c}{\multirow{3}[2]{*}{\textbf{Out-of-distribution}}} & \multicolumn{3}{c}{\textbf{AUROC}} & \multicolumn{3}{c}{\textbf{TNR@TPR95}} \\
    \cmidrule(lr){2-4} \cmidrule(lr){5-7}
          & \multicolumn{1}{c}{\textbf{ConvNet}} & \multicolumn{1}{c}{\textbf{VGG-16}} & \multicolumn{1}{c}{\textbf{ResNet-20}} & \multicolumn{1}{c}{\textbf{ConvNet}} & \multicolumn{1}{c}{\textbf{VGG-16}} & \multicolumn{1}{c}{\textbf{ResNet-20}} \\
          \cmidrule(lr){2-4} \cmidrule(lr){5-7}
          & \multicolumn{3}{c}{\textbf{G-ODIN/Siamese}} & \multicolumn{3}{c}{\textbf{G-ODIN/Siamese}} \\
    \midrule[0.25ex]
    \multicolumn{7}{c}{\textbf{blur}} \\
    \midrule
    \textbf{MNIST} & 69.50/69.29 & 60.88/85.15 & 78.00/78.38 & 13.67/50.11 & 02.00/56.23 & 45.37/48.25 \\
    \textbf{FMNIST} & 73.99/69.85 & 71.51/57.58 & 74.86/75.50 & 37.16/50.05 & 38.26/28.40 & 27.17/44.98 \\
    \textbf{CIFAR-10} & 77.09/79.97 & 45.28/85.28 & 73.70/76.51 & 52.69/55.70 & 04.44/36.95 & 40.96/39.94 \\
    \midrule
    \multicolumn{7}{c}{\textbf{bright}} \\
    \midrule
    \textbf{MNIST} & 58.77/73.70 & 60.23/81.83 & 89.00/88.11 & 21.66/49.96 & 38.30/46.43 & 57.01/67.57 \\
    \textbf{FMNIST} & 78.62/84.70 & 74.14/73.45 & 73.79/76.82 & 40.18/59.94 & 49.54/33.03 & 33.02/41.62 \\
    \textbf{CIFAR-10} & 77.04/95.39 & 44.77/86.90 & 77.00/82.71 & 55.20/79.80 & 04.34/37.69 & 46.65/60.06 \\
    \midrule
    \multicolumn{7}{c}{\textbf{contrast}} \\
    \midrule
    \textbf{MNIST} & 58.71/71.73 & 73.74/83.43 & 88.38/84.39 & 20.37/49.90 & 31.87/48.71 & 58.90/69.68 \\
    \textbf{FMNIST} & 74.82/81.98 & 71.77/70.30 & 79.33/75.79 & 38.70/57.32 & 41.12/31.23 & 30.39/41.46 \\
    \textbf{CIFAR-10} & 72.73/89.78 & 41.73/82.51 & 69.22/79.95 & 50.02/70.45 & 03.44/32.64 & 39.19/53.08 \\
    \midrule
    \multicolumn{7}{c}{\textbf{zoom}} \\
    \midrule
    \textbf{MNIST} & 67.43/85.15 & 66.85/77.19 & 75.24/82.75 & 03.97/54.38 & 03.82/43.96 & 32.92/56.41 \\
    \textbf{FMNIST} & 81.28/83.77 & 70.67/70.10 & 81.56/83.05 & 44.66/57.68 & 33.30/31.66 & 37.57/51.03 \\
    \textbf{CIFAR-10} & 88.25/94.35 & 47.60/90.11 & 88.92/91.11 & 64.49/74.78 & 05.87/40.93 & 53.71/56.82 \\
    \midrule
    \multicolumn{7}{c}{\textbf{Mean}} \\
    \hline
    \midrule
    
    \textbf{MNIST} & 63.61/\textbf{74.97} & 65.43/\textbf{81.90} & 82.66/\textbf{83.41} & 14.92/\textbf{51.09} & 18.99/\textbf{48.83} & 48.55/\textbf{60.48} \\
    \textbf{FMNIST} & 77.18/\textbf{80.08} & \textbf{72.02}/67.86 & 77.39/\textbf{77.79} & 40.17/\textbf{56.25} & \textbf{40.56}/31.08 & 32.04/\textbf{44.77} \\
    \textbf{CIFAR-10} & 78.78/\textbf{89.87} & 44.84/\textbf{86.20} & 77.21/\textbf{82.57} & 55.60/\textbf{70.18} & 04.52/\textbf{37.05} & 45.13/\textbf{52.48} \\
    \bottomrule[0.25ex]
    \end{tabular}%
\end{table*}%

Table \ref{tab:sia_tr} compares the performance of G-ODIN,
SelfChecker, and our distribution analyzer on detecting deviated data
from in-distribution data on three DL models and three datasets with
four transformations. Since SelfChecker is designed as a classifier,
we omit its TNR@TPR95 here.

Except for CIFAR-10 on ConvNet, the distribution analyzer 
outperforms G-ODIN. The reason it fails is that ConvNet is too
simple a DNN for CIFAR-10 on which Siamese cannot learn informative
embeddings. But G-ODIN decomposed the softmax score to distinguish
confident inputs from unconfident ones, which is hardly affected by the
model architecture~\cite{hsu2020generalized}.

In all other settings, Siamese achieves both high AUROC (averagely 91.84\% against 75.01\% of G-ODIN) and TNR@TPR95 (averagely 57.20\% against 29.76\% of G-ODIN)
that is harder to reach. That means that Siamese can recognize more
in-distribution data with the 95\% recall of deviated data, which is
important for the downstream reflection process. Given by the threshold search from validation dataset, Siamese can correctly detect on average 77.19\% of deviated data.
SelfChecker has bad
performance since it is constrained by the assumption that the testing
instance conform to the distribution of training dataset. But the
deviated data here share similar semantics with the training data but
are far away.

\begin{table*}[h]
	\centering
	\caption{Performance of three methods on detecting deviated data.}
	\label{tab:sia_tr}
    \begin{tabular}{lcccccc}
    \toprule[0.25ex]
    \multicolumn{1}{c}{\multirow{3}[2]{*}{\textbf{Deviated}}} & \multicolumn{3}{c}{\textbf{AUROC}} & \multicolumn{3}{c}{\textbf{TNR@TPR95}} \\
    \cmidrule(lr){2-4} \cmidrule(lr){5-7}
          & \textbf{ConvNet} & \textbf{VGG-16} & \textbf{ResNet-20} & \textbf{ConvNet} & \textbf{VGG-16} & \textbf{ResNet-20} \\
          \cmidrule(lr){2-4} \cmidrule(lr){5-7}
          & \multicolumn{3}{c}{\textbf{G-ODIN/SelfChecker/Siamese}} & \multicolumn{3}{c}{\textbf{G-ODIN/Siamese}} \\
    \midrule[0.25ex]
    \multicolumn{7}{c}{\textbf{blur}} \\
    \midrule
    \textbf{MNIST} & \ \ \ \ 74.62/64.47/99.96\ \ \ \ & \ \ \ \ 59.14/60.19/99.73\ \ \ \ & \ \ \ \ 71.01/70.10/99.81 \ \ \ \ & 19.21/99.99 & 06.61/55.37 & 24.33/93.92 \\
    \textbf{FMNIST} & 93.61/62.62/99.39 & 85.90/63.20/96.16 & 78.72/61.68/94.56 & 66.43/99.97 & 47.43/59.52 & 29.90/57.89 \\
    \textbf{CIFAR-10} & 90.28/64.44/90.61 & 52.47/64.91/86.16 & 82.20/64.86/88.35 & 47.37/32.41 & 06.12/21.97 & 21.27/37.26 \\
    \midrule
    \multicolumn{7}{c}{\textbf{bright}} \\
    \midrule
    \textbf{MNIST} & 99.95/61.03/99.99 & 50.14/57.40/99.61 & 58.08/66.37/97.48 & 100.0/99.96 & 03.47/66.31 & 06.29/86.96 \\
    \textbf{FMNIST} & 89.13/67.31/99.84 & 83.24/67.31/91.94 & 72.65/71.95/97.58 & 48.97/88.84 & 48.30/19.94 & 22.10/88.12 \\
    \textbf{CIFAR-10} & 88.86/60.78/75.23 & 57.69/51.24/73.00 & 73.95/66.28/82.13 & 45.95/18.08 & 06.55/13.23 & 15.02/26.65 \\
    \midrule
    \multicolumn{7}{c}{\textbf{contrast}} \\
    \midrule
    \textbf{MNIST} & 99.93/60.12/99.99 & 66.79/69.29/99.62 & 55.13/64.17/97.12 & 100.0/100.0 & 04.58/61.18 & 05.39/84.30 \\
    \textbf{FMNIST} & 90.82/54.90/99.90 & 76.06/45.69/95.51 & 71.76/63.97/97.77 & 55.98/90.46 & 26.93/29.09 & 24.01/88.41 \\
    \textbf{CIFAR-10} & 93.37/58.14/75.87 & 63.95/56.07/87.35 & 85.32/64.15/89.39 & 64.07/16.24 & 11.33/29.13 & 24.64/35.79 \\
    \midrule
    \multicolumn{7}{c}{\textbf{zoom}} \\
    \midrule
    \textbf{MNIST} & 83.33/45.87/99.67 & 68.41/44.76/98.47 & 88.42/64.15/97.42 & 30.39/98.91 & 04.00/71.80 & 51.83/86.69 \\
    \textbf{FMNIST} & 80.63/59.24/95.45 & 65.76/67.02/86.04 & 63.94/59.88/88.54 & 32.10/68.60 & 07.42/33.36 & 14.55/49.09 \\
    \textbf{CIFAR-10} & 80.34/49.88/73.46 & 42.54/45.82/76.82 & 62.34/60.67/76.42 & 31.56/15.05 & 03.79/17.27 & 13.55/17.20 \\
    \midrule
    \multicolumn{7}{c}{\textbf{Mean}} \\
    \hline
    \midrule
    \textbf{MNIST} & 89.46/57.87/\textbf{99.91} & 61.12/57.91/\textbf{99.36} & 68.16/66.20/\textbf{97.96} & 62.40/\textbf{99.72} & 04.67/\textbf{63.67} & \textbf{21.96}/\textbf{87.97} \\
    \textbf{FMNIST} & 88.55/61.02/\textbf{98.65} & 77.74/60.81/\textbf{92.41} & 71.77/64.37/\textbf{94.62 }& 50.87/\textbf{86.97} & 32.52/\textbf{35.48} & \textbf{22.64}/\textbf{70.88} \\
    \textbf{CIFAR-10} & \textbf{88.21}/58.31/78.79 & 54.16/54.51/\textbf{80.83} & 75.95/63.99/\textbf{84.07} & \textbf{47.24}/20.45 & \textbf{06.95}/20.40 & \textbf{18.62}/\textbf{29.23} \\
    \bottomrule[0.25ex]

    \end{tabular}
\end{table*}

\begin{figure}[htpb]
	\vspace{-1.1em}
	\includegraphics[width=.85\linewidth]{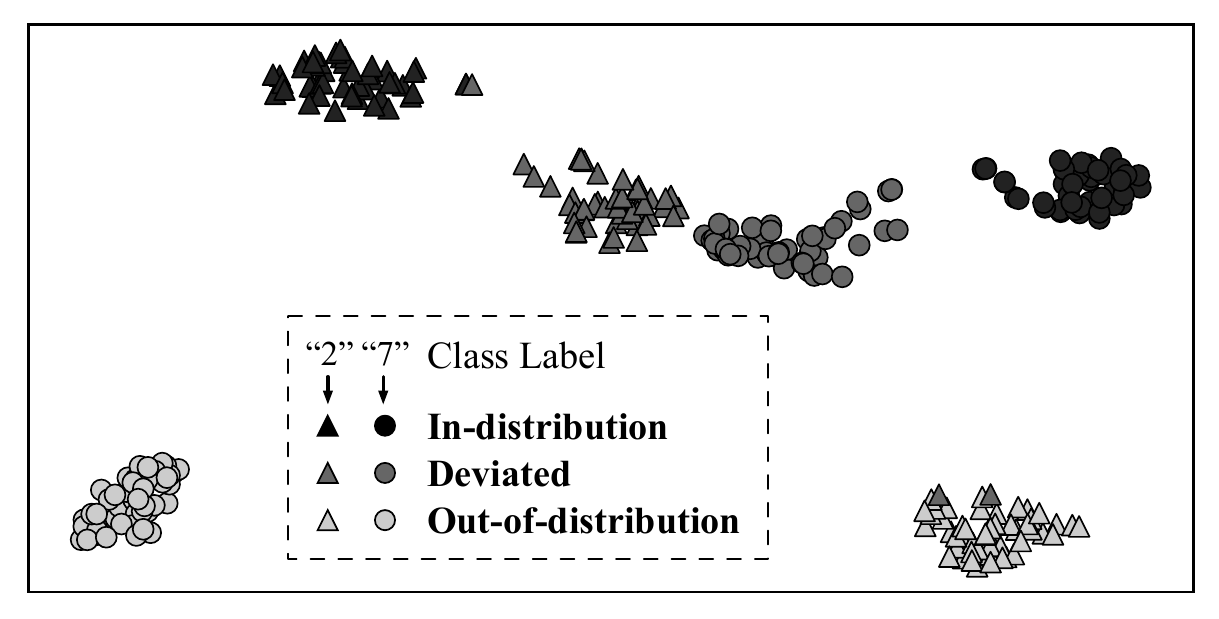}
	\par
	\caption{Visualization of the feature embedding learned by the Siamese network.}
	\label{fig:tsne}
	\vspace{-1.1em}
\end{figure}

Figure \ref{fig:tsne} visualizes the separation between three types of
data on MNIST with two labels ("2" and "7"). This is generated from the learned embeddings from Siamese
network and then mapped into hyperspace by t-SNE~\cite{van2008visualizing}. We can see that the three
types of data are distant from each other and that the distance
between in-distribution and deviated data is \emph{shorter} than the
distance between in-distribution and out-of-distribution data.

For RQ1, we conclude that the distribution analyzer has high
performance in distinguishing three kinds of data. It can effectively
detect out-of-distribution and deviated data with high accuracy.

\begin{figure*}[t]
	\centering\includegraphics[width=1.0\linewidth]{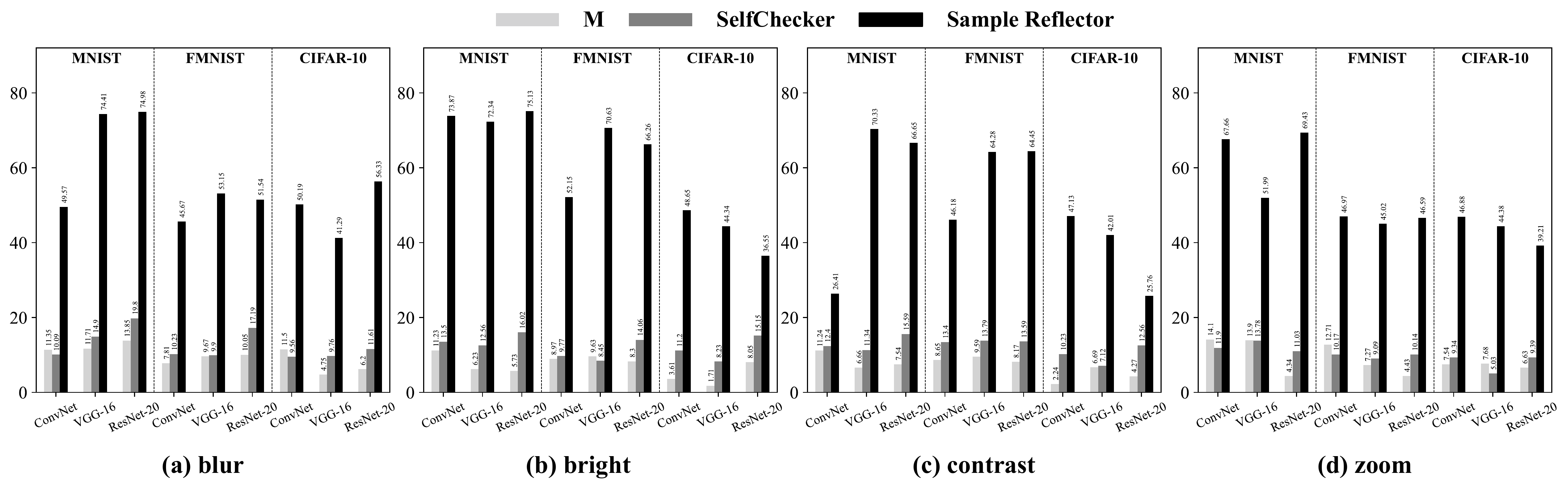}
	\par
	\caption{Accuracy of three methods on unseen deviated input.}
	\label{fig:reflect}
\end{figure*}

\vspace*{0.2em}\noindent\textbf{RQ2. Sample Reflector Accuracy.}\vspace*{0.2em}
Figure \ref{fig:reflect} compares the accuracies of the subject model
$M$, SelfChecker, and the sample reflector on unseen deviating
data. All four kinds of deviated data have not been seen by these
models.

We find that the sample reflector achieves higher
accuracy than $M$ and SelfChecker. The poor performance of SelfChecker
is predictable since the deviating data does not strictly conform to
the distribution of the training data.


For RQ2, we showed that the sample reflector prediction can improve
the accuracy of the subject models on unseen deviating data.

\vspace*{0.2em}\noindent\textbf{RQ3. \tool Performance.}\vspace*{0.2em}
To show the effectiveness of the proposed \tool, Table
\ref{tab:reflect} presents the accuracies of the subject model $M$,
$M$ with data augmentation ($M+Aug$), SelfChecker, and \tool on the
in-distribution and deviating testing data. Both $M+Aug$ and \tool use
the "blur" version of the training data. $M+Aug$ enhances the original
training data with its blurry version to retrain $M$. \tool uses
blurry training data to train the auxiliary models, i.e., Siamese and
Quadruplet network, instead of retraining M.

As we can see, SelfChecker cannot improve the $M$'s accuracy since it
has poor performance in giving alternative prediction for deviating
data. Both data augmentation (averagely 84.02\%) and \tool (averagely 84.63\%) improve the
accuracy of $M$ (averagely 50.08\%). The difference in their accuracies is not
significant, which shows that \tool has similar generalization ability as
data augmentation.

\begin{table*}
	\caption{Accuracy of four methods on the in-distribution and deviating testing data.}
	\label{tab:reflect}
	\begin{tabular}{m{3cm}m{3.2cm}m{3.2cm}  m{3cm}}
		\toprule[0.25ex]
		\multirow{2}[1]{*}{\textbf{Accuracy}} & \multicolumn{1}{c}{\textbf{ConvNet}} & \multicolumn{1}{c}{\textbf{VGG-16}} & \multicolumn{1}{c}{\textbf{ResNet-20}} \\
		\cmidrule(lr){2-4}

		& \multicolumn{3}{c}{\textbf{M/M+Aug/SelfChecker/InputReflector}} \\
		\midrule[0.25ex]
		\multicolumn{4}{c}{\textbf{blur}} \\
		\midrule
		\textbf{MNIST} & 55.16/99.38/57.19/99.36 & 55.65/99.55/60.45/99.02 & 56.65/99.44/55.39/99.40 \\
		\textbf{FMNIST} & 50.04/93.67/54.30/92.48 & 51.94/94.41/55.14/91.44 & 51.40/91.60/52.17/91.57 \\
		\textbf{CIFAR-10} & 45.84/78.43/47.29/76.08 & 46.85/88.59/44.39/88.42 & 43.91/79.50/50.78/79.20 \\
		\midrule
		\multicolumn{4}{c}{\textbf{bright}} \\
		\midrule
		\textbf{MNIST} & 55.10/93.14/56.09/92.11 & 52.91/96.49/59.81/95.22 & 52.59/91.64/60.37/94.43 \\
		\textbf{FMNIST} & 50.62/75.13/49.82/81.18 & 51.92/75.32/56.12/77.40 & 50.53/80.17/60.25/84.65 \\
		\textbf{CIFAR-10} & 41.89/64.88/47.94/65.23 & 45.33/70.99/48.45/77.43 & 44.84/65.73/46.86/65.14 \\
		\midrule
		\multicolumn{4}{c}{\textbf{contrast}} \\
		\midrule
		\textbf{MNIST} & 55.11/92.76/57.83/90.77 & 53.12/95.74/58.56/93.83 & 53.49/92.67/60.06/94.62 \\
		\textbf{FMNIST} & 50.46/75.73/59.26/80.09 & 51.90/76.30/58.30/77.10 & 50.46/80.69/57.95/86.66 \\
		\textbf{CIFAR-10} & 41.21/74.87/42.64/75.92 & 47.82/79.96/49.18/83.14 & 42.95/72.88/45.83/72.76 \\
		\midrule
		\multicolumn{4}{c}{\textbf{zoom}} \\
		\midrule
		\textbf{MNIST} & 56.54/90.52/57.23/85.92 & 56.74/84.90/56.38/84.87 & 51.89/83.32/59.43/90.37 \\
		\textbf{FMNIST} & 52.49/79.56/51.65/78.49 & 50.74/84.90/50.28/81.66 & 48.59/80.59/49.54/81.54 \\
		\midrule
		\textbf{CIFAR-10} & 43.86/76.41/49.25/77.06 & 48.32/86.39/50.03/84.09 & 44.13/78.38/53.56/78.26 \\
		
		\midrule
		\multicolumn{4}{c}{\textbf{Mean}} \\
		\hline
		\midrule
		\textbf{MNIST} & 55.47/\textbf{93.95}/57.09/92.04 & 54.60/\textbf{94.17}/58.80/93.23 & 53.65/91.77/58.81/\textbf{94.70} \\
		\textbf{FMNIST} & 50.90/81.02/53.76/\textbf{83.06} &51.62/\textbf{82.73}/54.96/81.90 & 50.24/83.26/54.98/\textbf{86.10} \\
		\textbf{CIFAR-10} & 43.20/\textbf{73.65}/46.78/73.57 & 47.08/81.48/48.01/\textbf{83.27} & 43.95/\textbf{74.12}/49.26/73.84\\
		\bottomrule[0.25ex]
	\end{tabular}
\end{table*}

However, not all data used in data augmentation can enhance $M$. For
example, the accuracy of $M$ will degrade when using adversarial
example for data augmentation.

To show the sensitivity of $M+Aug$ and \tool on adversarial/poisonous
data, we include both adversarial and blurry training data for both
methods. Table \ref{tab:adv} shows the mean accuracies of $M+Aug$ and
\tool on the in-distribution data and the one plus deviating data with
the four transformations.

The lowest accuracies of $M+Aug$ on in-distribution data indicate that
data augmentation with adversarial examples will harm model
performance even on the normal testing data. But, \tool is more
tolerant and it does not effect the accuracy of in-distribution
data. But, the generalization ability of both methods are reduced, as
indicated by the lower accuracies on in-distribution and deviating
data than when only using blurry training data as augmentation. The
tolerance of \tool benefits from the sample mining in both
Siamese and Quadruplet networks where many adversarial examples are
filtered out.

\begin{table*}[h]
	\centering
	\caption{Accuracy using adversarial examples as data augmentation.}
	\label{tab:adv}
	\begin{tabular}{lcccccc}

		\toprule[0.25ex]
		\multicolumn{1}{c}{\multirow{3}[2]{*}{\textbf{Accuracy}}} & \multicolumn{3}{c}{\textbf{In-distribution}} & \multicolumn{3}{c}{\textbf{In-distribution+Deviated}} \\
		\cmidrule(lr){2-4} \cmidrule(lr){5-7}

		& \textbf{ConvNet} & \textbf{VGG-16} & \textbf{ResNet-20} & \textbf{ConvNet} & \textbf{VGG-16} & \textbf{ResNet-20} \\
		\cmidrule(lr){2-4} \cmidrule(lr){5-7}
		& \multicolumn{3}{c}{\textbf{M/M+Aug/InputReflector}} & \multicolumn{3}{c}{\textbf{M/M+Aug/InputReflector}} \\
		
		\midrule[0.25ex]
		
		\textbf{MNIST}    & 98.97/\underline{97.46}/99.57                    & 99.58/\underline{97.54}/99.65                   & 99.44/\underline{98.31}/99.48  & 55.47/88.52/\textbf{89.20}                    & 54.60/90.67/\textbf{94.38 }                  & 53.65/93.00/\textbf{94.81  }                      \\
		\textbf{FMNIST}   & 92.27/\underline{92.08}/92.27                    & 94.20/\underline{92.20}/94.57                    & 92.75/\underline{91.34}/93.58 & 50.90/79.50/\textbf{82.03}                    & 51.62/80.38/\textbf{85.99 }                  & 50.24/80.89/\textbf{83.05 }   \\
			\textbf{CIFAR-10} & 80.17/\underline{76.80}/80.71                    & 88.95/\underline{87.59}/89.31                   & 81.62/\underline{75.38}/82.96  & 43.20/\textbf{68.73}/66.31                    & 47.08/76.35/\textbf{79.01 }                  & 43.95/64.39/\textbf{67.07 }\\
		\bottomrule[0.25ex]
	\end{tabular}
\end{table*}

To answer RQ3, \tool has similar generalization ability as data
augmentation and can detect out-of-distribution data but data augmentation doesn't have such mechanism.
And \tool is more tolerant than data augmentation when considering
adversarial examples.

\vspace*{0.2em}\noindent\textbf{RQ4. Overhead.}\vspace*{0.2em}
We measured the time consumption of $M+Aug$, SelfChecker, and \tool
during training and deployment. Table \ref{tab:time} shows that $M+Aug$ and \tool cost similar training time where SelfChecker costs the most since it needs to estimate the distribution of each layer. For each test instance in deployment,  \tool needs more time overhead than $M+Aug$ because of the distance calculation and closet training data search. But its time consumption is much lower than SelfChecker.

\begin{table}[h]
		\centering
	\caption{Time overhead.}
	\begin{tabular}{lccc}
		\toprule[0.25ex]
		\textbf{Time} & \textbf{M+Aug }& \textbf{Selfchecker} & \textbf{InputReflector} \\ \hline
		\textbf{Training}   &   41.97m    &     >1h        &        39.5m        \\ 
		\textbf{Deployment} &    0.98s   &     34.33s        &      2.86s          \\ \bottomrule[0.25ex]

	\end{tabular}
	\label{tab:time}
\end{table}

\section{Related Work}
\label{sec:related}

\tool contains two elements: a distribution analyzer to detect
out-of-distribution and deviated data, and a sample reflector. In this
section, we will discuss previous work related to both of these
elements.

\textbf{Detection of out-of-distribution data.}
Several methods \cite{liang2018enhancing, lee2018simple,
  hsu2020generalized} have been proposed to detect out-of-distribution
data that is completely different from the training data. ODIN
\cite{liang2018enhancing} and Generalized ODIN
\cite{hsu2020generalized} are based on trained neural network
classifiers that must be powerful enough for the specific dataset.  In
\cite{liang2018enhancing, hsu2020generalized}, the authors modified
the last softmax layer to statistically distinguish in-distribution
and out-of-distribution data. The max class probability of the softmax
layer are mapped into a hyperspace by temperature scaling and input
pre-processing to create a more effective score that can distinguish
the distributions of out-of-distribution and in-distribution data. Hsu
et al. \cite{hsu2020generalized} introduced a decomposed confidence
function to avoid using out-of-distribution data to train the
classifier, as was done by Liang et al. \cite{liang2018enhancing}. The
decomposed confidence function splits the original softmax function
into two parts to avoid issues with overconfidence.

The internal representation of DNN models given an input has also been
used to detect out-of-distribution data.  Mahalanobis
\cite{lee2018simple} takes hidden layers of the DNN model as
representation spaces. This work used the distance calculation and
input pre-processing to compute the Mahalanobis distance to measure the
extent to which an input belongs to the in-distribution in these
spaces. But there is a hyperparameter in input pre-processing that need
to be tuned for each out-of-distribution dataset.

None of these techniques detect deviated data that is between
in-distribution and out-of-distribution data, which is the key feature
of our \tool. Hsu et al. \cite{hsu2020generalized} showed that Generalized ODIN
is empirically superior to previous approaches
\cite{hendrycks2016baseline, liang2018enhancing, lee2018simple}; we therefore
use Generalized ODIN as our baseline to evaluate the performance of the
distribution analyzer in \tool.

\textbf{Data augmentation.}
To improve the DNN model generalization, data augmentation
techniques have been proposed~\cite{gao2020fuzz,
  samangouei2018defense,zhong2020random,dreossi2018counterexample}.
In a recent work in this space, Guo et al. \cite{gao2020fuzz} augment training
data using mutation-based fuzzing. The
authors specifically focus on improving the robust generalization of
DNNs. A model with robust generalization should not degrade in
performance for train/test data with small perturbations such as
applying spatial transformations to images). By drawing a parallel
between training DNNs and program synthesis, mutation-based fuzzing is
used for data augmentation and the computational cost has been further
reduced by selecting only a part of data for augmentation.
%
%
But, our experimental results indicate that DNN model accuracy on
normal testing data will quickly degrade if the training data is
augmented with data containing large perturbations. The aim of \tool
is therefore to improve model generalization without sacrificing its accuracy.
Another problem with data augmentation is that even an augmented training dataset
is ultimately finite, limiting model generalization.


\textbf{Runtime trustworthiness checking.}
In the software engineering community, several studies consider
checking a DNN's trustworthiness in deployment.  DISSECTOR
proposed by Wang et al. \cite{wang2020dissector} detects inputs that
deviate from normal inputs. It trains several sub-models on top of the
pre-trained model to validate samples fed into the
model. Xiao et al. \cite{xiao2021self} propose SelfChecker to monitor
DNN outputs using internal layer features. It trigger an alarm if the
internal layer features of the model are inconsistent with the final
prediction and also provides an alternative prediction. The assumption
in these two papers is that the training and validation datasets come
from a distribution similar to that of the inputs that the DNN model
will face in deployment.
For example, SelfChecker cannot detect out-of-distribution
data and provides an alarm constrained by this assumption.
 SelfChecker's
 ability of analyze unseen deviating data is also weak, not to
 mention provide predictions for such input.

\section{Conclusion}
\label{sec:conclusion}

Deployed DNNs must contend with inputs that may contain noise or
distribution shifts. Even the best-performing DNN model in training
may make wrong predictions on such inputs. 
%
We presented an input reflection approach to deal with this
issue. Input reflection first determines that the input is problematic
and then reflects it towards a nearby sample in the training
dataset. Unlike previous work, this reflection strategy focuses on
characteristics of the input and allows for fine-grained control and
better understanding of precisely when and how the technique works. We
implemented input reflection as part of the \tool tool and evaluated
it empirically across several datasets and model variants. On the
three popular image datasets with four transformations \tool is able to distinguish unseen problematic inputs
with an average accuracy of 77.19\%. Furthermore, by combining
\tool with the original DNN, we can increase the average model
prediction accuracy by 34.55\% on the in-distribution and deviated testing dataset. Moreover, \tool can detect out-of-distribution data that the original DNN cannot handle.
We hope that our work inspires
further research into characterizing the deployment-time inputs in the
context of the DNN training dataset.

\balance
\bibliographystyle{ACM-Reference-Format}
\bibliography{main}

\end{document}